\documentclass{egpubl}
\usepackage{pg2023}

\usepackage[T1]{fontenc}
\usepackage{dfadobe}  
\usepackage{amsmath}
\usepackage{amssymb}
\usepackage{times}
\usepackage{booktabs}
\usepackage{multicol}
\usepackage{tabularx}
\usepackage{color}
\usepackage{cite}
\BibtexOrBiblatex
\electronicVersion
\PrintedOrElectronic
\ifpdf \usepackage[pdftex]{graphicx} \pdfcompresslevel=9
\else \usepackage[dvips]{graphicx} \fi

\usepackage{egweblnk} 

\newcommand{\eg}{e.g. }
\newcommand{\ie}{i.e. }

\title[A Perceptual Shape Loss for Monocular 3D Face Reconstruction]%
      {A Perceptual Shape Loss for Monocular 3D Face Reconstruction}

\author[C. Otto et al.]{\parbox{\textwidth}{\centering
		\vspace{-5mm}
		C. Otto$^{1,2}$\orcid{0000-0002-5625-593X}, P. Chandran$^{1}$\orcid{0000-0001-6821-5815}, G. Zoss$^{1}$\orcid{0000-0002-0022-8203}, M. Gross$^{1,2}$\orcid{0000-0002-6838-9775}, P. Gotardo*$^{1}$\orcid{0000-0001-8217-5848}, D. Bradley$^{1}$\orcid{0000-0002-2055-9325}
		\bigbreak
\small{$^1$DisneyResearch|Studios, Switzerland\\
	$^2$ETH Zürich, Switzerland      }}
}

\begin{document}

\maketitle
\begin{abstract}
   Monocular 3D face reconstruction is a wide-spread topic, and existing approaches tackle the problem either through fast neural network inference or offline iterative reconstruction of face geometry. In either case carefully-designed energy functions are minimized, commonly including loss terms like a photometric loss, a landmark reprojection loss, and others.  In this work we propose a new loss function for monocular face capture, inspired by how humans would perceive the quality of a 3D face reconstruction given a particular image.  It is widely known that shading provides a strong indicator for 3D shape in the human visual system.  As such, our new `perceptual' shape loss aims to judge the quality of a 3D face estimate using only shading cues.  Our loss is implemented as a discriminator-style neural network that takes an input face image and a shaded render of the geometry estimate, and then predicts a score that perceptually evaluates how well the shaded render matches the given image. This `critic' network operates on the RGB image and geometry render alone, without requiring an estimate of the albedo or illumination in the scene.  Furthermore, our loss operates entirely in image space and is thus agnostic to mesh topology.  We show how our new perceptual shape loss can be combined with traditional energy terms for monocular 3D face optimization and deep neural network regression, improving upon current state-of-the-art results. \\

\ccsdesc[500]{Computing methodologies~Mesh geometry models}

\printccsdesc   
\end{abstract}  

\section{Introduction}
\label{sec:intro}
 \footnote[0]{*Now at Google}
Monocular 3D face reconstruction is a primary research problem in computer vision and provides the ability to recover 3D geometry and motion of a person's face from images or videos, and opens up many applications in entertainment, telepresence and interaction in virtual environments.  This problem has been studied for a long time, and many methods and benchmarks have been proposed~\cite{nowbenchmark,realy}.  Methods for monocular 3D face reconstruction range from inference-based approaches that focus on training a neural network to predict the face shape from an image to others that directly optimize for the face shape given an image.  

In either case, the one common theme is that most methods formulate an energy minimization problem (either during training, or during optimization) with loss terms that aim to make the resulting geometry match the shape of the person's face in the image. Common loss terms include a pixel-wise photo-consistency loss, a landmark alignment loss, and sometimes more specialized loss terms like identity or emotion recognition losses. In this work we propose a new loss term based on human perception of shape that can be added to a wide variety of techniques used in monocular face capture.

Our work is motivated by the human perception of shape and our ability to use shading cues to estimate 3D orientations of surfaces in images. For example, Ramachandran~\cite{Ramachandran1988} showed that humans can perceive shape from shading using intensity alone, assuming the object is lit from a single light source. Based on this inspiration, we design a new neural loss function that takes an RGB image of a face and a shaded render of the geometry estimate as input, and predicts a scalar value (\eg a score) to indicate how well the geometry matches the shape of the face, including identity, pose and expression.  Following the findings of Ramachandran~\cite{Ramachandran1988}, we render the geometry estimate with diffuse gray shading, lit from a single light source (\eg using PyTorch3D~\cite{pytorch3d}). We implement our perceptual shape loss as a discriminator (or {\em critic}) network, based on the Deep Convolutional GAN (DCGAN) architecture~\cite{dcgan}, and train it with both real and fake image-render pairs, obtained from a dataset of 3D facial scans in a studio setting augmented with in-the-wild synthetic data.

Our critic network operates entirely in image space and it is agnostic of the mesh topology, which is beneficial since different algorithms rely on different 3D face models, like the Basel face model (BFM)~\cite{bfm2,bfm}, FLAME~\cite{flame}, FaceWarehouse~\cite{facewarehouse} or other custom topology meshes~\cite{facescape,hifi3d,lsfm,lyhm,realy,faceverse,wood2,facedb}. As such, our trained critic network can be easily used as a loss term in energy functions for 3D face reconstruction.  

We demonstrate the use of our perceptual shape loss in the scenario of offline optimization-based monocular 3D face reconstruction as well as in the scenario of deep neural network 3D face regression. Specifically, in an optimization framework we show that the addition of our loss improves overall reconstruction accuracy as compared to optimizing a standard set of loss functions used in literature, when evaluated using the NoW benchmark~\cite{nowbenchmark}.  Similarly, in an inference framework we show that using our loss to train a new 3D face regression method (which we call PSL) by fine-tuning DECA~\cite{deca}, improves the results of identity and expression as judged by both the NoW benchmark~\cite{nowbenchmark} and the REALY benchmark~\cite{realy}.  To summarize:

\begin{itemize}
\item We present a new perceptual shape loss function for monocular face reconstruction, inspired by the human ability to distinguish shape from shading. 
\item Our loss function is topology-agnostic, and can be used with any 3D face model. 
\item Our loss is generic, and can be readily incorporated into energy functions used for 3D face reconstruction from images in both optimization and regression frameworks. 
\item We train a novel Perceptual Shape Loss (PSL) method for 3D face regression by fine-tuning DECA with our loss function and show that its reconstructions improve identity and expression accuracy on the NoW and the REALY benchmarks.\\
\end{itemize}

\section{Related Work}
\label{sec:related_work}

High-quality 3D face reconstructions have traditionally required expensive multi-camera studio setups~\cite{beeleretal}. Great research effort has gone into exploring alternative approaches that can alleviate many of the traditional constraints (\eg multiple cameras, controlled lighting, expensive hardware) by investigating single-camera, in-the-wild 3D face reconstruction. Most monocular methods use a 3D Morphable Model (3DMM)~\cite{blanz} as a statistical face prior for reconstruction, to account for ambiguities, such as the missing depth information~\cite{Wu2016}. 3DMMs are created from a corpus of 3D face scans, and rely on statistics to represent plausible human face shapes and expressions in a data-driven way.  Detailed surveys of 3DMMs and their use in monocular face capture are given in Egger et al.~\cite{3dmmssota}, Zollhöfer et al.~\cite{3drecsota}, and Morales et al.~\cite{morales}. Most state-of-the-art 3D face reconstruction methods are based on either the FLAME face model~\cite{flame} or the BFM~\cite{bfm,bfm2}.
The best 3DMM shape, expression and pose parameters for a given RGB image are searched either via analysis-by-synthesis optimization~\cite{blanz,garrido,dib_optimization,ganfit,face2face,thies_opt} or via deep neural network regression~\cite{hrn,albedogan,mofa,tran1,genova,richardson,richardson2,inversefacenet,reg_method,cnnregression,tewari_reg}.

One way to evaluate the quality of state-of-the-art methods is to compare their 3D face reconstructions with the corresponding ground truth face scans. The {\textit{not quite in-the-wild}} (NoW) benchmark~\cite{nowbenchmark} and the benchmark proposed by Feng et al.~\cite{fengbenchmark} evaluate the quality of the neutral face shape (\ie the identity). The REALY benchmark \cite{realy,lyhm,facescape,hifi3d} evaluates both identity and expression, because their ground truth face scans also include non-neutral faces. Both NoW and REALY follow ICP-based protocols to align the face reconstructions with ground truth face scans, establish correspondences, and compute vertex errors. Two of the best ranking methods on the NoW benchmark to date are MICA~\cite{mica} and DECA~\cite{deca}, both of which are based on the FLAME face model. Further state-of-the-art methods as judged by the NoW benchmark are either BFM-based ~\cite{focus,dib,synergynet,3ddfav2,deng,mgcnet,3dmmcnn}, FLAME-based~\cite{nowbenchmark,pymafx2022}, based on other face models~\cite{wood,umdfa} or are model-free~\cite{prnet}. Most of these methods use traditional loss functions when training inference networks, like pixel-wise photo-consistency and landmark alignment. Our proposed perceptual loss function could readily complement the majority of these approaches.

EMOCA~\cite{emoca} and SPECTRE~\cite{spectre} are two recent methods that introduce new perceptual loss terms to improve upon the quality of the reconstructions predicted by DECA. EMOCA uses an emotion CNN trained on an annotated large scale emotion dataset~\cite{affectnet} to recognize emotions in RGB images and rendered reconstructions. They compare the predicted features in the last network layer, effectively creating a perceptual loss by enforcing consistency between the two. SPECTRE improves upon the lip reconstruction by running a pre-trained lip-reading network~\cite{lipreadingnetwork} for detecting features from lip RGB image crops and rendered lip reconstructions. Similar to these two works, we also utilize a CNN
 to arrive at our perceptual shape judgements. In contrast to both methods, our perceptual shape loss maximizes a global score based on a shaded render input and the target image
 , drawing inspiration from the human ability of leveraging shading cues to perceive shape~\cite{Ramachandran1988}.  In terms of quality, the reconstructions by EMOCA display more faithful emotions than previous methods, and SPECTRE improves upon the visual quality of the lip reconstruction. However, the visual improvements of both methods are reflected only in certain categories of the REALY benchmark~\cite{realy} and not reflected in the NoW benchmark~\cite{nowbenchmark}. One reason for this is that both methods do not improve upon the identity component initialized by DECA, focusing only on the expression or jaw pose components. On the other hand, the visual improvements gained with our perceptual shape loss are also reflected by quantitative improvements in the identity and the expression reconstruction, which we measure on both the NoW and the REALY benchmarks (Section~\ref{sec:reconstruction}).
\section{Perceptual Shape Loss}
\label{sec:method}

Our key idea is to create a loss function for monocular 3D face reconstruction, implemented as a neural network that takes a face image and a gray-shaded render of face geometry as input, and outputs a scalar value to indicate how well the render matches the image in terms of shape. The network should intuitively critique the inputs, and provide continuous feedback about the `goodness' of match between the image and the render, for any image-render pair. Once such a critic network is trained, its output can be interpreted as a perceptual shape loss that can be used for 3D face reconstruction tasks. In the following, we describe our network design, training data, and other implementation details.

\subsection{Critic Network}
\label{subsec:criticNet}

\begin{figure}[ht]
\center
\includegraphics[width=1\columnwidth]{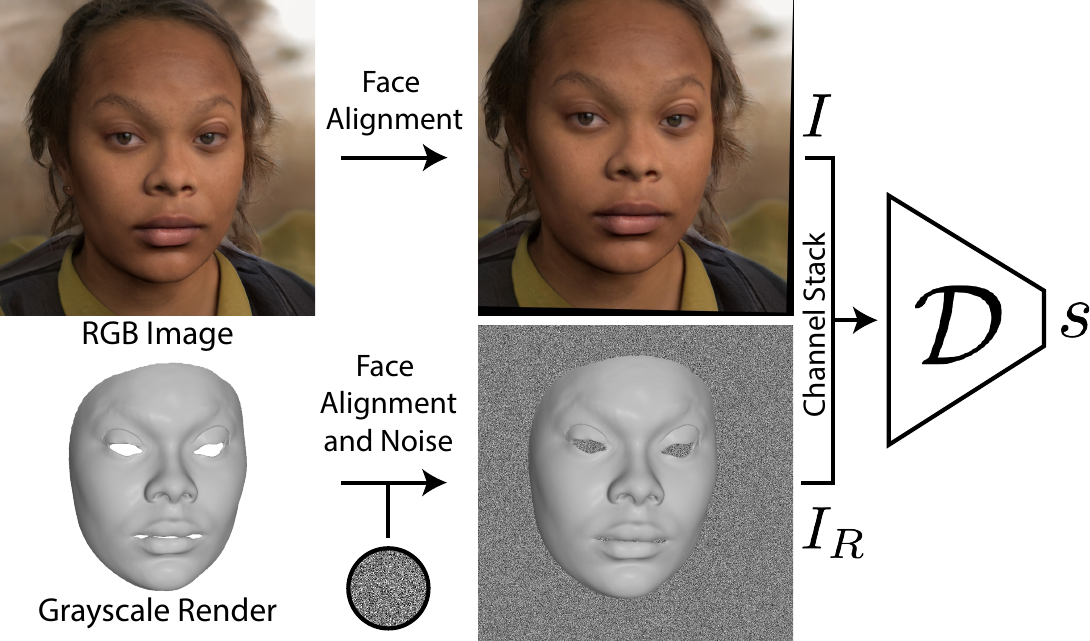}
\caption{Our convolutional network $\mathcal{D}$ takes an aligned RGB face image and a gray-shaded geometry render as input, and predicts a score $\mathcal{S}$ that indicates how well the render matches the image in terms of shape.}
\label{fig:networkOverview}
\end{figure}

Our network is implemented as a CNN, inspired by the Deep Convolutional GAN (DCGAN)~\cite{dcgan} discriminator. The DCGAN discriminator is a simple and well-explored architecture that is commonly used for training critic neural networks~\cite{wgangp,wgan}. It consists only of several strided convolution layers, normalization layers and leaky ReLU activation functions and is therefore fast to train. There is no activation function after the last layer, allowing for a real valued scalar output. To accommodate for our input resolution of $256 \times 256$ we extend the original DCGAN (which only supported $64 \times 64$ images) by two additional convolution layers. The input to the network is an RGB face image $I$ and a single channel gray-shaded render of a geometry estimate $I_R$, rendered from the same camera viewpoint and shaded by a single point light in front of the face.  The image $I$ and the render $I_R$ are stacked to make a 4-channel input which is fed to the critic $\mathcal{D}$.  The output of our critic network is a 1-dimensional scalar $\mathcal{S}$, which will represent the perceptual shape loss value.  An overview of the network is shown in Figure \ref{fig:networkOverview}.  

We train the network as a discriminator to regress higher values when the image-render pair match well, and lower values when they match poorly.  Our training data includes a set of {\em real} image-render pairs where the render matches the image perfectly, and a set of {\em fake} image-render pairs where the render does not match the image (see Section \ref{subsec:trainingData} for more details on the training data).  We formulate our training objective so that the score for real examples is maximized and for fake examples it is minimized, using the WGAN-GP loss~\cite{wgangp,wgan}:

\begin{equation}
	\mathcal{L}_{\text{critic}} = \mathcal{D}\left(\mathbf{\tilde{x}}\right) - \mathcal{D}\left(\mathbf{x}\right) +  \lambda\left(\|\nabla_\mathbf{\hat{x}}\mathcal{D}\left(\mathbf{\hat{x}}\right)\|_2- 1\right)^2,
\end{equation}

\noindent where $\mathcal{D}$ is our network, $\mathbf{x}$ is a real sample and $\mathbf{\tilde{x}}$ is a fake sample. A gradient penalty term encourages the gradient norm $\|\nabla_\mathbf{\hat{x}}\mathcal{D}\left(\mathbf{\hat{x}}\right)\|_2$ to stay close to 1 (1-Lipschitz constraint), where $\mathbf{\hat{x} = \epsilon x + (1 - \epsilon) \tilde{x}}$ and $\mathbf{\epsilon}$ is a random number drawn from ${U[0,1]}$~\cite{wgangp}. 
The loss $\mathcal{L}_{\text{critic}}$ is calculated during training after passing both a  batch of real and fake examples through the critic network.
Although our critic outputs a continuous real valued scalar during training, we re-normalize its output at inference time such that the median of real examples is $1.0$ and median of fake examples is $-1.0$, to have an intuitive loss.

\subsection{Training Data}
\label{subsec:trainingData}

For our training dataset, we use high-quality 3D face scans acquired from a multi-view studio capture setup~\cite{facedb}.  The dataset contains 358 different identities, displaying 24 different expressions under 12 different camera viewpoints.  The geometry is reconstructed using the passive stereo method of Beeler et al.~\cite{beeleretal}, and fit to a common template topology~\cite{facedb}.

\begin{figure}[t]
\center
\includegraphics[width=1\columnwidth]{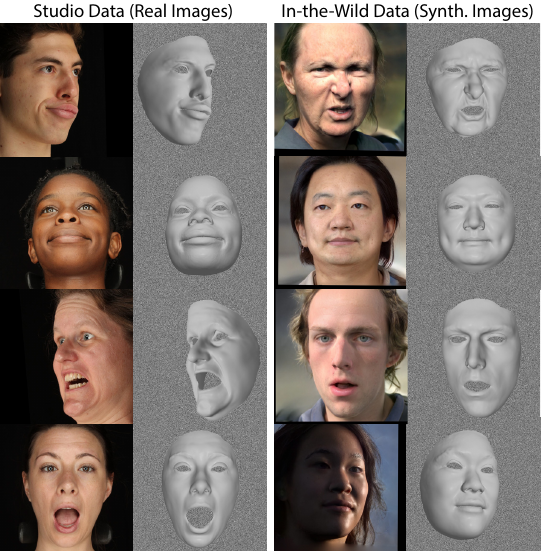}
\caption{Ground truth `real' samples we use for training.  Our dataset includes both studio images of real people (left) and in-the-wild synthetic images (right).}
\label{fig:trainingDataReal}
\end{figure}

\begin{figure}[t]
\center
\includegraphics[width=1\columnwidth]{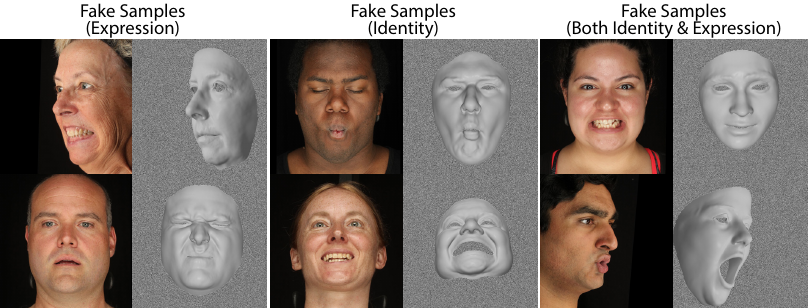}
\caption{We create `fake' samples for training by intentionally mixing face images with geometry renders that have incorrect expression, identity or both.  Additional fake images are created with incorrect head pose (not shown in the figure).}
\label{fig:trainingDataFake}
\end{figure}

To create real samples $\mathbf{x}$ we can select from the identities, expressions and cameras and use the corresponding reconstructed geometry to create the render, as it perfectly matches the identity and expression in the face skin area (excluding eyes, inner mouth and hair).  Some examples of these real samples used for training are shown in Figure \ref{fig:trainingDataReal} (left side).  To create fake examples $\mathbf{\tilde{x}}$ we purposely select mismatching images and geometry renders (\eg an image of one identity but the geometry render of a different identity).  Examples of such fake samples for training are illustrated in Figure \ref{fig:trainingDataFake}, where the fake render may have an incorrect expression, identity or both.

One problem with the studio dataset is that all the images are acquired from a fixed set of cameras, under uniform white illumination.  To encourage closure of the domain gap between studio and in-the-wild data, we create additional data using the approach of Chandran et al.~\cite{neuralfacedb}, who render the textured geometry in various scenes and inpaint the eyes, hair and background using a 2D human face image generator (StyleGAN2~\cite{stylegan2}).  Their approach allows control over viewpoint and lighting to imitate in-the-wild conditions. The resulting dataset contains a large variety of quasi in-the-wild images for which we have the corresponding high-quality geometry serving as ground truth.  Given this synthetic data, we can create real and fake samples using the same approach as for studio data.  Some real examples from this in-the-wild synthetic dataset are shown in Figure \ref{fig:trainingDataReal} (right side).  Note that the approach of Chandran et al.~\cite{neuralfacedb} is limited to mostly frontal viewpoints and does not allow for the creation of extreme side pose examples.

In total, we train our network on 92736 real studio examples and 6276 real synthetic examples. We generate three times the number of fake examples in the following way: One third of the fake examples have renders displaying an incorrect identity, another third display an incorrect expression and the last third display both, incorrect identity and expression.  To encourage our loss to critique the head pose, we create another 99012 fake examples where we apply spurious head rotations and translations to the geometry before rendering. This leads to a total of 495060 unique training examples (approximately 100K real and 400K fake). During training we oversample the real examples by a factor of 4 to equalize the number of real and fake examples seen by the network. We create an additional 20736 examples (10368 real and 10368 fake) to serve as a validation set.

\subsection{Implementation Details}

Prior to training we normalize the 4D input ($I$, $I_R$) images by cropping the faces and aligning the eyes, mouth and nose to approximately the same position across all inputs~\cite{swapshop}. As we aim to operate on any face topology including both full head geometry as well as frontal face meshes only, we ensure a consistent render area by discarding vertices at a distance greater than \textit{0.7 × (outer eye dist + nose dist)} from the center of the face, as suggested by related methods~\cite{nowbenchmark,fengbenchmark}, before rendering the geometry.  We further crop away any vertices from the interior of the mouth and eye region that exist in the geometry from Chandran et al.~\cite{facedb}, because many methods do not reconstruct these. For our training dataset, the final cropped mesh topology consists of 38799 vertices.

We follow Gulrajani et al.~\cite{wgangp} and do not use batch normalization in our critic network and replace it with instance normalization~\cite{instance_norm}. We implement our method in PyTorch~\cite{pytorch}.  We train our critic network for $4$ epochs using the Adam optimizer~\cite{adam} with a learning rate of $0.001$ and $\lambda=10$ on a TITAN X GPU. Training takes around eight hours (2 hours per epoch) with input images of size $256^2$ and batch size $64$ to reach the best separation between fake and real data on a validation set (refer to Section \ref{sec:validation} and Figure \ref{fig:histogramRealFake}). We create our renders with PyTorch3D~\cite{pytorch3d}.  We apply several color transformations to the RGB face images using Kornia data augmentations~\cite{kornia_one,kornia_two} in order to increase the variance of our dataset. Similar to McDonagh et al.~\cite{white_noise}, we use uniform noise as the background of our render so that the model learns the signal only from the foreground.

\section{Validation}
\label{sec:validation}

\begin{figure}[t]
\center
\includegraphics[width=1\columnwidth]{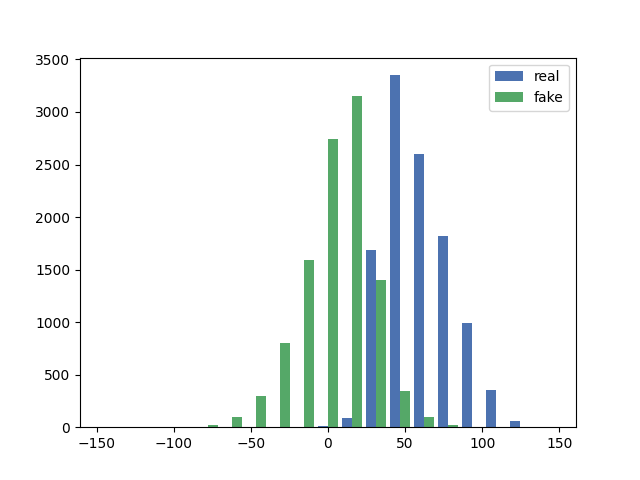}
\caption{Histogram of score separation on our held-out validation set. Note: These show the scores prior to $-1$ to $1$ normalization.}
\label{fig:histogramRealFake}
\end{figure}

We now present experiments to validate our critic network as a perceptual shape loss.  We first validate that our network learned to identify real versus fake input examples by plotting the distribution of output scores for our held-out validation set.  As illustrated in Figure \ref{fig:histogramRealFake}, our network has learned a reasonable separation of the two distributions.  Note that we plot the original output scores $\mathcal{S}$ before normalizing the medians to $-1$ to $1$.

\begin{figure}[t]
\center
\includegraphics[width=1\columnwidth]{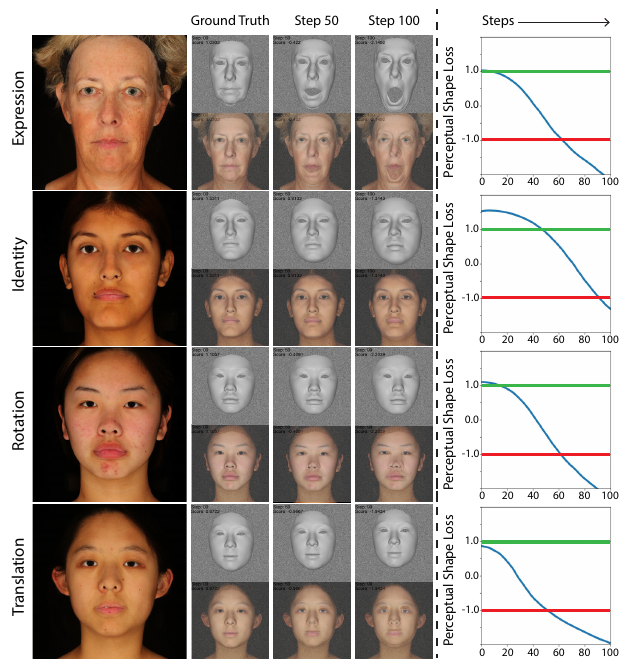}
\caption{We demonstrate the smooth deterioration of the perceptual shape score predicted by our critic network as the input shape gradually moves away from the ground truth. We show examples of the critic's response to deviation in expression, identity, rotation, and translation on four validation examples that were not seen by the critic during training. The smooth and intuitive degradation of the critic's score highlights the appropriateness of our perceptual shape loss for face reconstruction.}
\label{fig:validation}
\end{figure}

Equally important, we need to validate that our critic network is suitable for use as a loss function in gradient-based optimizations.  A loss function should be differentiable, continuous, smooth, and ideally have a minimum at the optimal parameter inputs.  As our perceptual loss is implemented as a CNN in PyTorch, it is trivially differentiable.  In order to demonstrate continuity and smoothness %
of the function, we perform a series of experiments where we plot the normalized output score while perturbing the geometry render in various ways.  Specifically, starting with the ground truth (real) render, we slowly edit the expression of the geometry so that it becomes increasingly farther from the ground truth. Figure \ref{fig:validation} (top row) shows several steps of this experiment, along with a plot of the per-step network output values.  We repeat this experiment three more times on different example images, this time varying the identity, head rotation and head translation, respectively (rows 2 to 4 in Figure \ref{fig:validation}).  As the plots indicate, our function is smooth and continuous, and generally decreases as we deviate from the optimal parameters. While not performing an exhaustive evaluation of the local neighborhood around the optimal parameters to show a strict maximum, our experiments show evidence of a local maximum for the optimal geometry renders. 

Given these successful validation experiments, we hypothesize that our critic network is suitable to use as a loss function.

\section{Face Reconstruction}
\label{sec:reconstruction}

\begin{figure}[ht]
	\center
	\includegraphics[width=1\columnwidth]{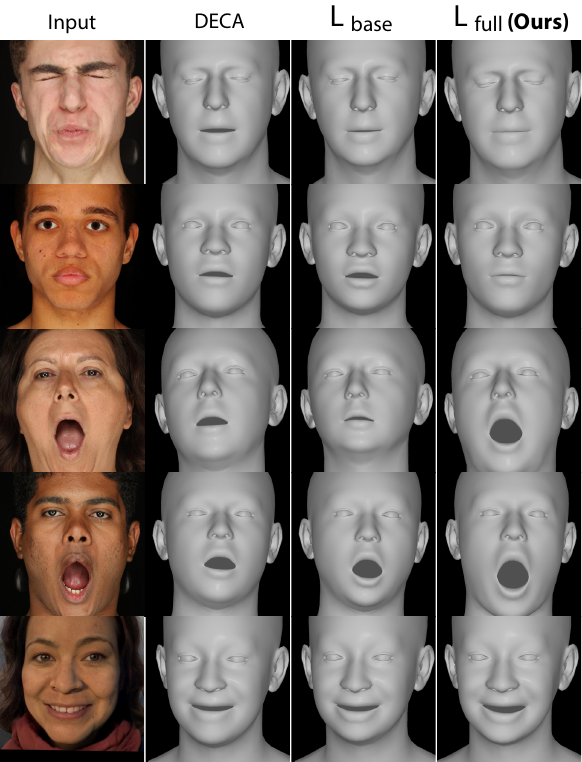}
	\caption{Here we show qualitative optimization results for face reconstruction on our held-out validation set, where the inferred geometry from DECA~\cite{deca} fails to faithfully reproduce the expressions seen in the input image. Even when this result is further optimized upon using traditional losses ($\mathcal{L}_{\text{base}}$), the visual quality of the result does not improve substantially. In contrast, the addition of the proposed perceptual loss ($\mathcal{L}_{\text{full}}$) enables the optimization to faithfully reproduce the expressions seen in the input image.}
	\label{fig:results_overview}
\end{figure}

We now demonstrate the application of our perceptual shape loss in the context of monocular 3D face reconstruction.  At a high level, the scenario is the following: given a face image and a user-defined deformable face model, we want to optimize or regress the model parameters in order to best match the pose and shape of the face in the image.  
Whether optimizing the single image directly (Section \ref{subsec:optimization}) or training a neural network for regression (Section \ref{subsec:regression}), our perceptual loss can be used by simply rendering the evaluated face geometry at each iteration and passing it through our critic network. For these experiments we will use the FLAME~\cite{flame} face model, and we will compare different energy terms, with and without our new loss function. 

The FLAME model consists of $\boldsymbol{V} \in \mathbb{R}\textsuperscript{5023$\times$3}$ vertices and 9976 faces $\boldsymbol{F}$, which are evaluated by the model function $\mathcal{M}$, as

\begin{equation}
	\mathcal{M( \boldsymbol{\beta},\boldsymbol{\theta},\boldsymbol{\psi})} \rightarrow(\boldsymbol{V},\boldsymbol{F}),
\end{equation}

\noindent where $\boldsymbol{\beta}$, $\boldsymbol{\psi}$ and $\boldsymbol{\theta}$ are the identity (shape), expression and pose parameters, respectively.  

As a framework for both the single-image optimization and the network training, we consider a set of standard energy terms defined by a recent regression-based face reconstruction method, DECA~\cite{deca}. 
Specifically, we formulate the following baseline energy function to minimize:

\begin{equation}
	\mathcal{L}_{\text{base}} = \lambda_{\text{l}}\mathcal{L}_{\text{l}} + \lambda_{\text{p}}\mathcal{L}_{\text{p}} + \lambda_{\text{id}}\mathcal{L}_{\text{id}} + \lambda_{\text{r}}\mathcal{L}_{\text{r}},
\end{equation}

\noindent where $\mathcal{L}_{\text{l}}$, $\mathcal{L}_{\text{p}}$, $\mathcal{L}_{\text{id}}$ and $\mathcal{L}_{\text{r}}$ are the landmark reprojection loss, photometric loss, identity loss, and regularization term defined in DECA.  As a comparison, we define a second energy function as

\begin{equation}
\mathcal{L}_{\text{full}} = \mathcal{L}_{\text{base}} + \lambda_{\text{s}}\mathcal{L}_{\text{s}},
\end{equation}

\noindent where $\mathcal{L}_{\text{s}}$ is our new perceptual shape loss that uses our pre-trained critic network $\mathcal{D}$, and is defined as

\begin{equation}
	\mathcal{L}_{\text{s}} = - D(I, I_R)
\end{equation}

\noindent given the RGB face image $I$ and the estimated geometry render $I_R$. We further demonstrate the power of our perceptual loss term by defining a third, more compact loss function as

\begin{equation}
	\mathcal{L}_{\text{compact}} = \lambda_{\text{l}}\mathcal{L}_{\text{l}} + \lambda_{\text{s}}\mathcal{L}_{\text{s}} + 
	\lambda_{\text{r}}\mathcal{L}_{\text{r}},
\end{equation}

\noindent which contains only our perceptual loss, the landmark reprojection loss and the  regularization losses, dropping the photometric and identity losses.  We keep the landmark loss because, like most face reconstruction methods, we rely on landmarks to robustly ground the reconstruction and achieve stable results over a wide range of images. In the following sections we evaluate the quality of different face reconstruction methods using the three loss functions $\mathcal{L}_{\text{base}}$, $\mathcal{L}_{\text{full}}$ and $\mathcal{L}_{\text{compact}}$.

\begin{table*}[h!]
	\centering
	\begin{multicols}{2}
		\scalebox{0.742}{
			\begin{tabularx}{1.335\textwidth}{l c c c c c}
				\toprule
				\textbf{NoW val. - Optimization} & \textbf{Full (Med/Mean/Std)$\downarrow$} & \textbf{Neutral (Med/Mean/Std)$\downarrow$} & \textbf{Expression (Med/Mean/Std)$\downarrow$} & \textbf{Occlusion (Med/Mean/Std)$\downarrow$} & \textbf{Selfie (Med/Mean/Std)$\downarrow$}\\
				\midrule
				Initialization (DECA) & 1.1757 / 1.4611 / 1.2511 & 1.1546 / 1.4433 / 1.2408 & 1.1351 / 1.4251 / 1.2384 & 1.2181 / 1.4890 / 1.2528 & 1.2562 / 1.5445 / 1.2996\\
				$\mathcal{L}_{\text{base}}$
				& 1.1241 / 1.4033 / 1.2062 & \textbf{1.1029} / \textbf{1.3749} / 1.1843 & 1.0700 / 1.3456 / 1.1788 & \textbf{1.1634} / \textbf{1.4403} / \textbf{1.2125} & 1.2566 / 1.5505 / 1.2990\\
				$\mathcal{L}_{\text{compact}}$ (Ours)
				& 1.1228 / 1.3987 / 1.1936 & 1.1046 / 1.376 / \textbf{1.1741} & 1.0699 / 1.3441 / 1.1678 & 1.1870 / 1.4673 / 1.2318 & \textbf{1.1901} / 1.4649 / 1.2235\\
				$\mathcal{L}_{\text{full}}$ (Ours)
				& \textbf{1.1226} / \textbf{1.3983} / \textbf{1.1932} & 1.1067 / 1.3778 / 1.1747 & \textbf{1.0694} / \textbf{1.3432} / \textbf{1.1665} & 1.1831 / 1.4650 / 1.2327 & 1.1907 / \textbf{1.4637} / \textbf{1.2209} \\
				\midrule
				\textbf{NoW val. - Inference-based} & \textbf{Full (Med/Mean/Std)$\downarrow$} & \textbf{Neutral (Med/Mean/Std)$\downarrow$} & \textbf{Expression (Med/Mean/Std)$\downarrow$} & \textbf{Occlusion (Med/Mean/Std)$\downarrow$} & \textbf{Selfie (Med/Mean/Std)$\downarrow$}\\
				\midrule
				EMOCA \cite{emoca} & 1.1757 / 1.4611 / 1.2511 & 1.1546 / 1.4433 / 1.2408 & 1.1351 / 1.4251 / 1.2384 & 1.2181 / 1.4890 / 1.2528 & 1.2562 / 1.5445 / 1.2996\\
				EMOCA\_V2 \cite{emoca} & 1.1757 / 1.4611 / 1.2511 & 1.1546 / 1.4433 / 1.2408 & 1.1351 / 1.4251 / 1.2384 & 1.2181 / 1.4890 / 1.2528 & 1.2562 / 1.5445 / 1.2996\\
				SPECTRE \cite{spectre} & 1.1757 / 1.4611 / 1.2511 & 1.1546 / 1.4433 / 1.2408 & 1.1351 / 1.4251 / 1.2384 & 1.2181 / 1.4890 / 1.2528 & 1.2562 / 1.5445 / 1.2996\\	
				DECA \cite{deca} & 1.1757 / 1.4611 / 1.2511 & 1.1546 / 1.4433 / 1.2408 & 1.1351 / 1.4251 / 1.2384 & 1.2181 / 1.4890 / 1.2528 & 1.2562 / 1.5445 / 1.2996\\
				$\mathcal{L}_{\text{base}}$
				& 1.0805 / \textbf{1.3489} / 1.1616 & 1.0544 / 1.3226 / 1.1463 & 1.0442 / 1.3162 / \textbf{1.1553} & \textbf{1.1300} / \textbf{1.3861} / \textbf{1.1647} & 1.1497 / 1.4267 / 1.2027 \\
				PSL - $\mathcal{L}_{\text{compact}}$ (Ours)
				& \textbf{1.0780} / 1.3538 / \textbf{1.1588} & 1.0454 / 1.3234 / \textbf{1.1418} & 1.0455 / 1.3289 /1.1558 & 1.1717 / 1.4266 / 1.1782 & \textbf{1.0707} / \textbf{1.3546} / \textbf{1.1645} \\
				PSL - $\mathcal{L}_{\text{full}}$ (Ours)
				& 1.0812 / 1.3573 / 1.1712 & \textbf{1.0440} / \textbf{1.3204} / 1.1488 & \textbf{1.0335} / \textbf{1.3154} / 1.1601 & 1.1850 / 1.4436 / 1.1970 & 1.1119 / 1.3920 / 1.1943 \\
				\bottomrule
		\end{tabularx}}
	\end{multicols}
	\vspace{1mm}
	\caption{\label{tab:full_now_eval_table}The NoW validation set results displaying the full score and the scores when evaluated on each of the four sub-challenges in isolation (multiview neutral, multiview expression, multiview occlusion and selfie). The NoW benchmark evaluates only the identity component of the reconstruction (no expression). For the inference-based comparison, we also compare our method with other state-of-the-art methods that fine-tune the DECA regression network (i.e. EMOCA \cite{emoca} and SPECTRE \cite{spectre}). As EMOCA \cite{emoca} and SPECTRE \cite{spectre} use the DECA \cite{deca} identity parameters, they achieve the same score as DECA. Errors are reported in mm.} 
\end{table*}

\begin{figure}[h!]
	\center
	\includegraphics[width=1\columnwidth]{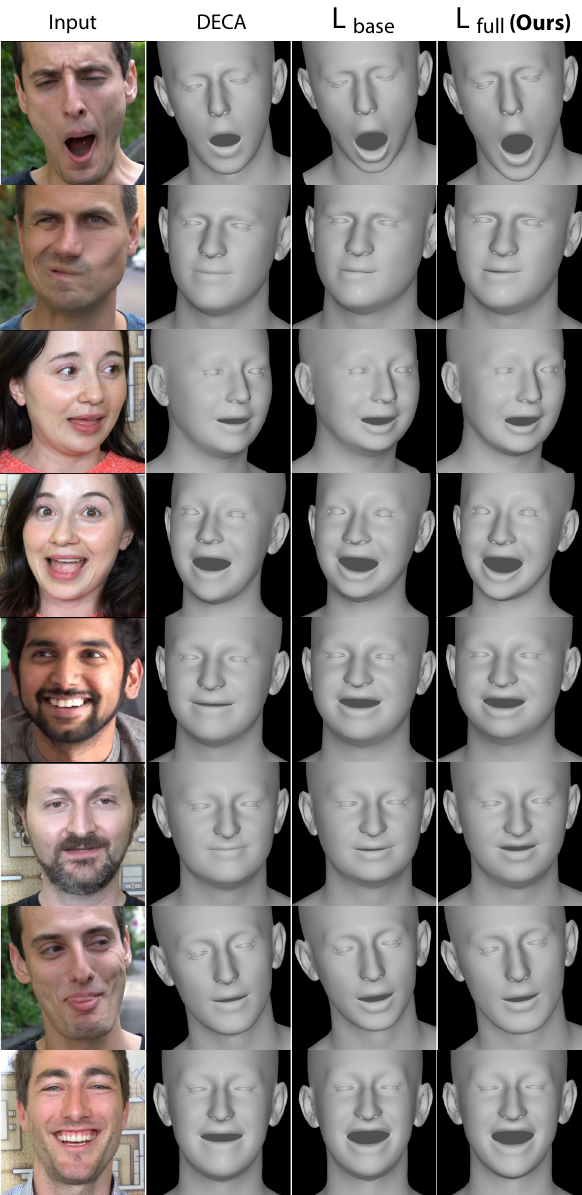}
	\caption{Here we show additional qualitative optimization results for face reconstructions on images in the wild, comparing the initial DECA~\cite{deca} fit with optimizing parameters without ($\mathcal{L}_{\text{base}}$) and with ($\mathcal{L}_{\text{full}}$) our perceptual loss.}
	\label{fig:results_overview2}
\end{figure}

\subsection{Single-Image Optimization}
\label{subsec:optimization}

In the context of offline single-image face reconstruction, we compare reconstruction accuracy while using each of the three loss functions during parameter optimization.  For evaluation, we use the common non-metrical NoW benchmark~\cite{nowbenchmark} and reconstruct both the NoW validation set and the NoW test set.  For every optimization, we initialize the model parameters using DECA's pre-trained coarse inference network and perform 60 optimization iterations with the following hyper-parameters: $\lambda_{\text{l}}=0.8$, $\lambda_{\text{p}}=2.0$, $\lambda_{\text{id}}=0.2$, $\lambda_{\text{r}}=0.0001$, and $\lambda_{\text{s}}=0.5$. Following DECA, we optimize for the first 100 identity parameters $\boldsymbol{\beta} \in \mathbb{R}\textsuperscript{100}$ and the first 50 expression parameters $\boldsymbol{\psi} \in \mathbb{R}\textsuperscript{50}$. We also optimize for a subset of $\boldsymbol{\theta} \in \mathbb{R}\textsuperscript{9}$ pose parameters describing global rotation, jaw joint and translation. The face albedo and illumination parameters required for the photometric and identity losses are fixed after initialization.

Optimization results on the validation set are shown in the upper section of Table \ref{tab:full_now_eval_table}, where we can see that using our new loss function ($\mathcal{L}_{\text{full}}$) increases the accuracy of reconstruction over the baseline. The greatest improvement over the baseline can be seen on the NoW selfie challenge subset, which contains mostly frontal faces. Note that all optimizations also naturally improve over the initialization (DECA), which is an inference-based method. Results on the NoW test set show similar improvements (upper section of Table \ref{tab:now_val_scores_nn}). Qualitative comparisons are also illustrated in Figure \ref{fig:results_overview} on our own held-out validation set of studio and in-the-wild data, where we can see that using our perceptual loss greatly helps to match the expression in the face image.  Additional optimization results are shown on an image dataset in-the-wild in Figure \ref{fig:results_overview2}.

We perform the validation and test set optimizations again using the compact loss function $\mathcal{L}_{\text{compact}}$. As we can see in Table \ref{tab:full_now_eval_table}, using only this compact loss can achieve favorable results, better than the baseline and nearly as good as the full set of loss terms. Note, that we choose to set $\lambda_{\text{r}}=0.0$ for $\mathcal{L}_{\text{compact}}$ during optimization, while still obtaining good results. This shows that our perceptual loss term is quite powerful on its own, and could replace several traditional losses.

\begin{table}[h!]
	\centering
	\begin{tabular}{l c c c}
		\toprule
		\textbf{NoW test - Optimization} & \textbf{Median $\downarrow$} & \textbf{Mean $\downarrow$} & \textbf{Std $\downarrow$}  \\
		\midrule
		Initialization (DECA) & 1.0936 & 1.3815 & 1.1829 \\
		$\mathcal{L}_{\text{base}}$
		& 1.0865 &  1.3760 &  1.1780 \\
		$\mathcal{L}_{\text{compact}}$ (Ours)
		& 1.0825 & 1.3700 & \textbf{1.1724} \\
		$\mathcal{L}_{\text{full}}$ (Ours)
		& \textbf{1.0818} & \textbf{1.3693} & \textbf{1.1724} \\
		\midrule
		\textbf{NoW test - Inference-based} & \textbf{Median $\downarrow$} & \textbf{Mean $\downarrow$} & \textbf{Std $\downarrow$}  \\
		\midrule
		EMOCA \cite{emoca} & 1.0936 & 1.3815 & 1.1829 \\
		SPECTRE \cite{spectre} & 1.0936 & 1.3815 & 1.1829 \\
		DECA \cite{deca} & 1.0936 & 1.3815 & 1.1829 \\
		$\mathcal{L}_{\text{base}}$
		& \textbf{1.0662} &  \textbf{1.3432} &  \textbf{1.1405} \\
		PSL - $\mathcal{L}_{\text{compact}}$ (Ours)
		& 1.0890 & 1.3680 & 1.1477 \\
		PSL - $\mathcal{L}_{\text{full}}$ (Ours)
		& 1.0902 & 1.3661 & 1.1453 \\
		\bottomrule
	\end{tabular}
	\vspace{1mm}
	\caption{\label{tab:now_val_scores_nn}Results on the non-metrical NoW benchmark test set \cite{nowbenchmark} of the initial DECA reconstruction versus our different loss terms for optimization (in mm). The optimization leads to best results when including our perceptual shape loss ($\mathcal{L}_{\text{full}}$), and even using our shape loss and landmark loss alone achieves good results ($\mathcal{L}_{\text{compact}}$). Similarly, we compare our the fine-tuned DECA regression network methods \cite{nowbenchmark} to the initial DECA reconstruction and other fine-tuned DECA regression networks (i.e. EMOCA \cite{emoca} and SPECTRE \cite{spectre}). While $\mathcal{L}_{\text{base}}$ performs best, PSL - $\mathcal{L}_{\text{full}}$ and PSL - $\mathcal{L}_{\text{compact}}$ also improve over the initialization.} 
\end{table}

\begin{table*}[h!]
	\centering
	\begin{multicols}{2}
	\scalebox{0.94}{
	\begin{tabularx}{1.048\textwidth}{l c c c c c}
		\toprule
		\textbf{REALY frontal-view} & \textbf{All (avg)$\downarrow$} & \textbf{Nose (avg/med/std)$\downarrow$} & \textbf{Mouth (avg/med/std)$\downarrow$} & \textbf{Forehead (avg/med/std)$\downarrow$} & \textbf{Cheek (avg/med/std)$\downarrow$} \\
		\midrule
		EMOCA \cite{emoca} & 2.103 & 1.868 / 1.821	
		 / 0.387 & 2.679 / 2.419 / 1.112 & 2.426 / 2.383 / 0.641 & \textbf{1.438} / \textbf{1.294} / \textbf{0.501}\\
		 EMOCA\_V2 \cite{emoca} & 2.230 & 1.704 / 1.680	
		 / 0.411 & 3.290 / 3.190 / 0.741 & \textbf{2.228} / \textbf{2.160} / 0.504 & 1.698 / 1.577 / 0.612\\
		 SPECTRE \cite{spectre} & 2.218 & 2.027 / 2.015	
		 / 0.401 & 2.695 / 2.623 / 0.969 & 2.335 / 2.263 / 0.581 & 1.817 / 1.703 / 0.585\\
		DECA \cite{deca} & 2.010 & 1.697 / 1.654 / 0.355 & 2.516 / 2.465 / 0.839 & 2.394 / 2.256 / 0.576 & 1.479 / 1.400 / 0.535\\
		$\mathcal{L}_{\text{base}}$
		& 2.223 & 1.804 / 1.768 / 0.434 & 2.899 / 2.801 / 0.937 & 2.391 / 2.321 / \textbf{0.535} & 1.796 / 1.715 / 0.564\\
		PSL - $\mathcal{L}_{\text{compact}}$ (Ours)
		& \textbf{1.840} & \textbf{1.658} / \textbf{1.591} / 0.353 & \textbf{1.832} / \textbf{1.775} / \textbf{0.549} & 2.394 / 2.395 / 0.552 & 1.475 / 1.386 / 0.515\\
		PSL - $\mathcal{L}_{\text{full}}$ (Ours)
		& 1.882 & 1.708 / 1.688 / \textbf{0.349} & 1.876 / 1.777 / 0.563 & 2.350 / 2.343 / 0.551 & 1.593 / 1.482 / 0.540 \\
		\midrule
		\textbf{REALY side-view} & \textbf{All (avg)$\downarrow$} & \textbf{Nose (avg/med/std)$\downarrow$} & \textbf{Mouth (avg/med/std)$\downarrow$} & \textbf{Forehead (avg/med/std)$\downarrow$} & \textbf{Cheek (avg/med/std)$\downarrow$} \\
		\midrule
		EMOCA \cite{emoca} & 2.125 & 1.867 / \textbf{1.548}	
		/ 0.554 & 2.636 / \textbf{1.738} / 1.284 & 2.448 / 2.369 / 0.708 & 1.548 / 1.424 / 0.590\\
		EMOCA\_V2 \cite{emoca} & 2.301 & 1.849 / 1.713	
		/ 0.726 & 3.283 / 3.238 / 0.830 & \textbf{2.316} / \textbf{2.256} / 0.653 & 1.758 / 1.621 / 0.669\\
		SPECTRE \cite{spectre} & 2.578 & 2.334 / 1.941	
		/ 1.322 & 3.210 / 2.839 / 1.467 & 2.567 / 2.433 / 0.725 & 2.203 / 1.885 / 1.580\\
		DECA \cite{deca} & 2.107 & 1.903 / 1.700 / 1.050 & 2.472 / 2.348 / 1.079 & 2.423 / 2.308 / 0.720 & 1.630 	
		/ 1.456 / 1.135 \\
		$\mathcal{L}_{\text{base}}$
		& 2.170 & 1.847 / 1.731 / 0.683 & 2.579 / 2.466 / 1.083 & 2.465 / 2.426 / 0.638 & 1.789 / 1.675 / 0.744 \\
		PSL - $\mathcal{L}_{\text{compact}}$ (Ours)
		& 1.931 & 1.715 / 1.621 / 0.591 & 1.989 / 1.836 / 0.786 & 2.535 / 2.492 / 0.634 & 1.486 / \textbf{1.358} / 0.592 \\
		PSL - $\mathcal{L}_{\text{full}}$ (Ours)
		& \textbf{1.857} & \textbf{1.685} / 1.629 / \textbf{0.475} & \textbf{1.820} / 1.757 / \textbf{0.557} & 2.454 / 2.426 / \textbf{0.608} & \textbf{1.469} / 1.378 / \textbf{0.495} \\
		\bottomrule
	\end{tabularx}}
	\end{multicols}
	\vspace{1mm}
	\caption{\label{tab:realy}Comparing different DECA-based neural network parameter regression methods on the REALY single image frontal-view and side-view benchmarks \cite{realy}. For both categories, fine-tuning DECA \cite{deca} with our perceptual shape loss (PSL) shows the lowest average error (`All' column) among the compared methods. We can observe large improvements especially in the mouth region, indicating high-quality expressions. Methods that fine-tune DECA with other perceptual loss terms (i.e. EMOCA \cite{emoca} and SPECTRE \cite{spectre}) do not perform as well. The numbers report the mean NMSE in mm.} 
\end{table*}

\subsection{Regression Network Training}
\label{subsec:regression}

\begin{figure*}[h!]
	\center
	\includegraphics[width=1\textwidth]{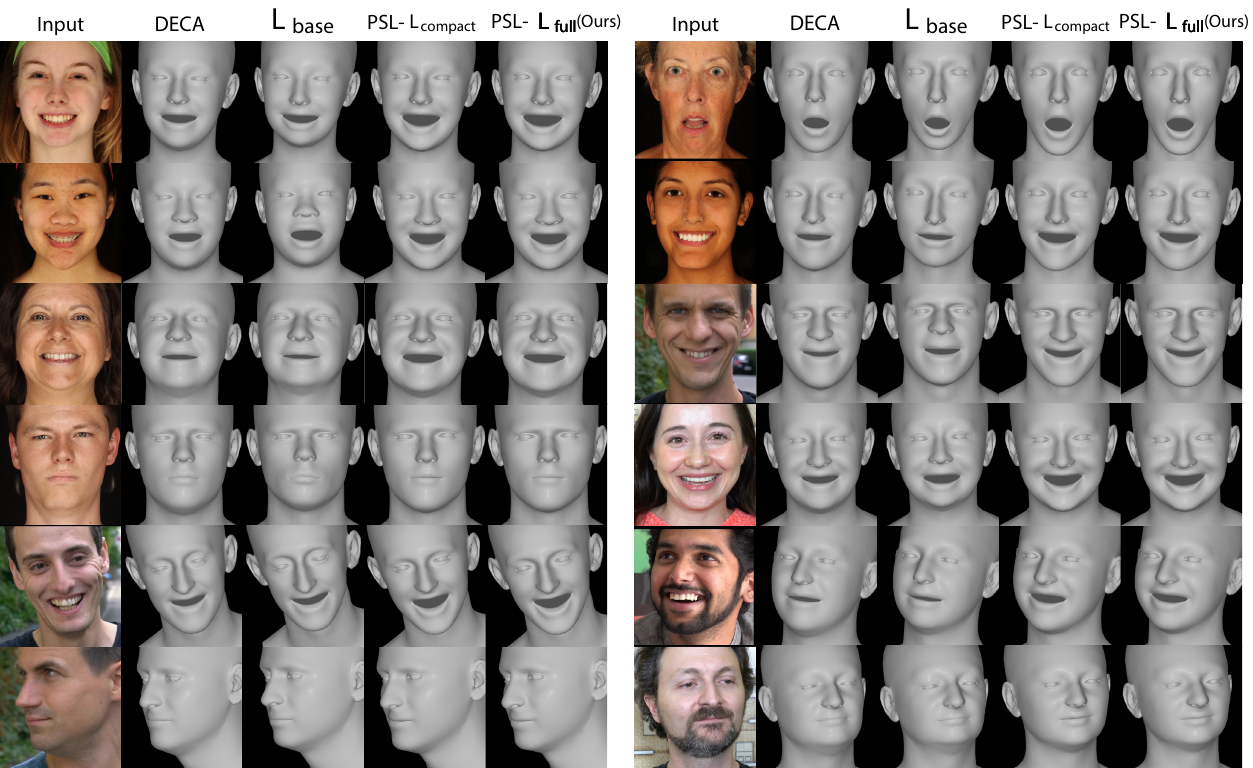}
	\caption{Here we show qualitative inference-based results for face reconstructions, comparing the initial DECA~\cite{deca} fit with $\mathcal{L}_{\text{base}}$ and the results achieved when using our perceptual shape loss during neural network training: PSL - $\mathcal{L}_{\text{compact}}$ and PSL - $\mathcal{L}_{\text{full}}$ .}
	\label{fig:loss-compare-regression}
\end{figure*}

\begin{figure*}[h!]
	\center
	\includegraphics[width=1\textwidth]{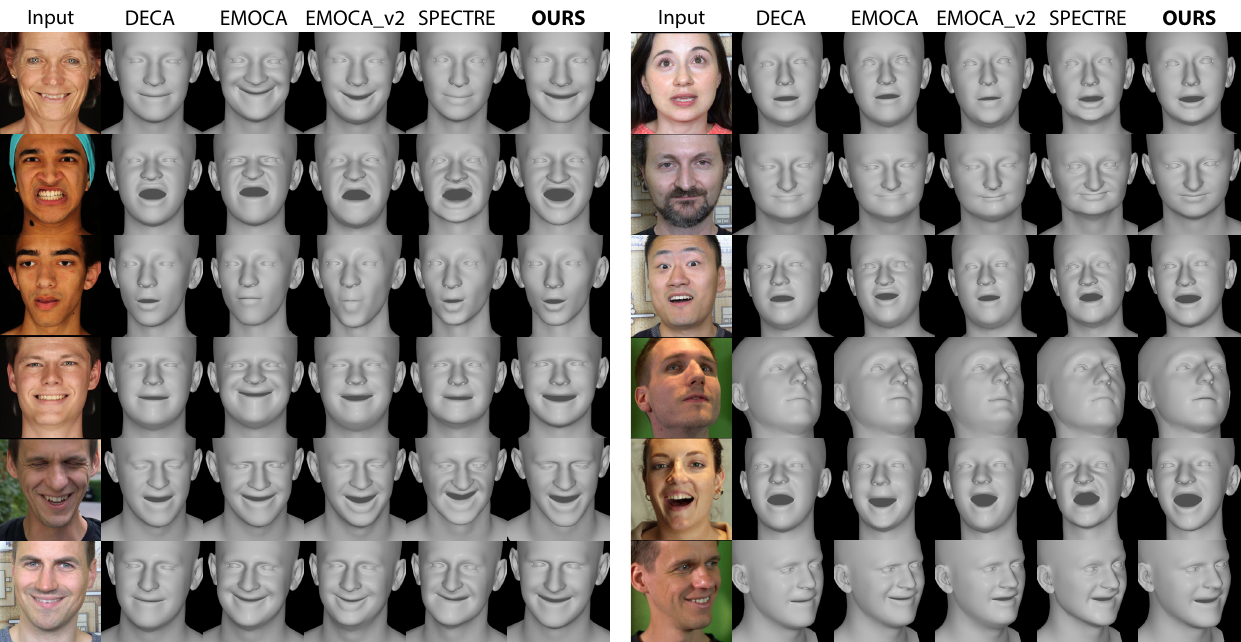}
	\caption{Here we show qualitative inference-based results for face reconstructions, comparing the initial DECA~\cite{deca} fit with our new method PSL - $\mathcal{L}_{\text{full}}$ and related state-of-the-art methods EMOCA~\cite{emoca}, EMOCA\_V2~\cite{emoca} and SPECTRE~\cite{spectre}.}
	\label{fig:method-compare-regression}
\end{figure*}

We now use our perceptual shape loss to train a new inference-based parameter regression method, based on the network architecture of DECA~\cite{deca}, which we call PSL (for \emph{perceptual shape loss}).  We train two versions of PSL - one using $\mathcal{L}_{\text{full}}$ and one using $\mathcal{L}_{\text{compact}}$. Additionally, we train a baseline inference network using $\mathcal{L}_{\text{base}}$.  Similar to related work~\cite{emoca,spectre} our training proceeds by fine-tuning a pre-trained DECA network. 

The optimization experiments in Section~\ref{subsec:optimization} show that our perceptual shape loss improves the most on frontal poses (Table~\ref{tab:full_now_eval_table} - Selfie) and expressions (Figure~\ref{fig:results_overview}).
Therefore, we fine-tune DECA using 2000 frontal images extracted from the CelebAMask-HQ dataset \cite{CelebAMaskhq}. We use a learning rate of $0.00001$ with the Adam optimizer~\cite{adam} and train for up to 1700 iterations. We set the following hyper-parameters: $\lambda_{\text{l}}=1.0$, $\lambda_{\text{p}}=2.0$, $\lambda_{\text{id}}=0.2$, $\lambda_{\text{r}}=0.0001$, and $\lambda_{\text{s}}=0.01$.

Once training is completed we evaluate the results quantitatively on the NoW benchmark to investigate identity improvement. We additionally evaluate PSL on the REALY benchmark \cite{realy}, which evaluates both identity and expressions. We compare models trained with $\mathcal{L}_{\text{base}}$, $\mathcal{L}_{\text{full}}$ and $\mathcal{L}_{\text{compact}}$. We show qualitative results produced by the different loss settings in Figure \ref{fig:loss-compare-regression}. We also compare with the state-of-the-art inference-based methods EMOCA \cite{emoca} and SPECTRE \cite{spectre}. Note that while other approaches (\eg MICA~\cite{mica} or HRN~\cite{hrn}) may show better overall performance on the benchmarks, we focus our comparison on EMOCA and SPECTRE because, similar to us, both methods fine-tune DECA and make use of their own perceptual losses during fine-tuning.

The results in the lower part of Table \ref{tab:full_now_eval_table} (Inference-based) show that using our perceptual shape loss improves upon the DECA initialization in terms of identity. We observe, similar to the optimization case, that the $\mathcal{L}_{\text{full}}$ and $\mathcal{L}_{\text{compact}}$ settings show the greatest improvement over $\mathcal{L}_{\text{base}}$ on the NoW Challenge selfie category. On the NoW test set (Table \ref{tab:now_val_scores_nn}), $\mathcal{L}_{\text{base}}$ achieves the best score. However, both $\mathcal{L}_{\text{full}}$ and $\mathcal{L}_{\text{compact}}$ still improve upon the DECA initialization. In contrast, EMOCA and SPECTRE keep the DECA identity parameters fixed and do not improve upon the DECA results on the NoW benchmark.  

Our PSL methods show large improvements on the REALY benchmark, when compared to other DECA-based parameter regression methods. Table \ref{tab:realy} displays the results for both the frontal-view and the side-view evaluation categories. In both categories, PSL-based networks outperform $\mathcal{L}_{\text{base}}$, EMOCA and SPECTRE. Among the compared methods, PSL - $\mathcal{L}_{\text{compact}}$ reaches the lowest error for the frontal-view category and PSL - $\mathcal{L}_{\text{full}}$ for the side-view category (`All' column). Furthermore, the results show large quantitative improvements in the mouth area for PSL-based methods, indicating high-quality expressions. These improvements can also be observed when looking at the qualitative results in Figure \ref{fig:method-compare-regression}.

\noindent\textbf{Summary:} 
In this section we have shown that our perceptual shape loss can be used in the context of monocular face reconstruction, and that employing our loss term in optimization or neural network training scenarios improves the accuracy and visual quality of the results.

\subsection{Topology Independence}
\label{subsec:topoIndependence}

\begin{figure}[h!]
	\centering
	\includegraphics[width=\columnwidth]{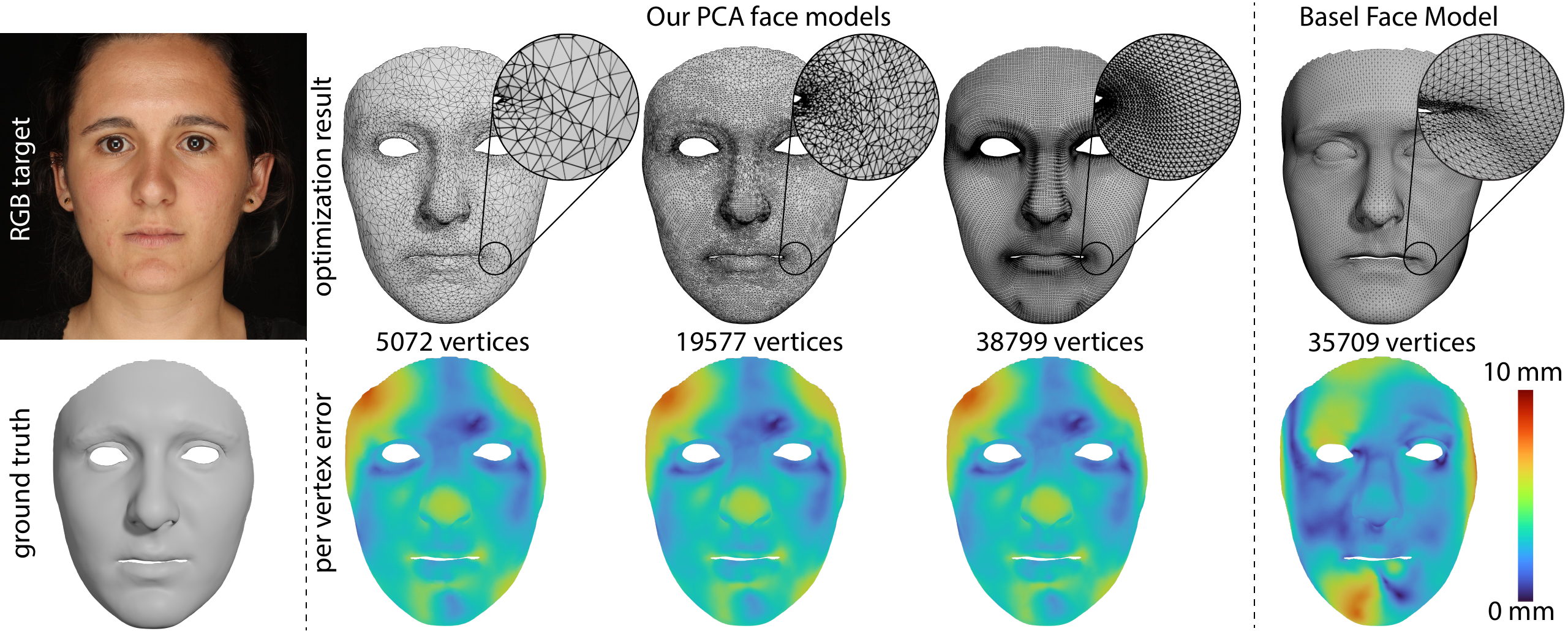}
	\caption{We employ our loss function on four different mesh topologies (5072, 19577, 35709 and 38799 vertices). Specifically, on our three PCA face models and the Basel Face Model (BFM)~\cite{bfm} used in Deep3D~\cite{deng}. The results of monocular face capture with our PCA face models show negligible differences across topologies. The BFM result differs slightly due to the different shape manifold of this model.}
	\label{fig:topo_independence}
\end{figure}

As our critic network operates only on images and shaded geometry renders it is agnostic to the actual face mesh topology. This means that the loss function is not tied to any particular deformable model, vertex count or UV layout.  To highlight this, we introduce an experiment where we use our perceptual loss on four different mesh topologies and show that our perceptual shape loss can be applied independent of the topology.

The experiment is again in the domain of optimization-based monocular face capture, similar to Section \ref{subsec:optimization}.  Here we isolate a single face image with neutral expression for illustration, and instead of using the FLAME face model we first create three separate PCA face models at three different resolutions: 5072 vertices, 19577 vertices, and 38799 vertices (refer to Figure \ref{fig:topo_independence}).  The PCA bases are created from the 3D reconstructions of neutral faces from the dataset of Chandran et al.~\cite{facedb}, after resampling the data to the aforementioned vertex counts.

For the sake of this experiment, we optimize for the parameters of each of the three PCA models using only our perceptual loss function $\mathcal{L}_{\text{s}}$, starting from the mean PCA face shape. Figure \ref{fig:topo_independence} shows that each optimization of our PCA face models produces the same result, with nearly identical error maps when compared to the ground truth shape scan. 

Additionally, we use our perceptual shape loss function $\mathcal{L}_{\text{s}}$ to optimize parameters of the Basel Face Model (BFM)~\cite{bfm} used in Deep3D~\cite{deng}, which includes expression parameters from  FaceWarehouse~\cite{facewarehouse,expbases}. Its topology consists of 35709 vertices and it operates in a different parameter space than the FLAME model or each of our PCA face models. We initialize the optimization from the mean BFM face with shape and expression parameters set to zero. Naturally, optimizing in the BFM parameter space leads to slightly different results when compared to the PCA face models, due to the different shape manifolds of each model. However, the optimization succeeds to capture the shape of the face in the image (Figure \ref{fig:topo_independence}).

These experiments indicate that our loss function is topology-agnostic, and our critic network does not need to be retrained when the topology (or model) changes. Further evidence can be found by the fact that the experiments in Section \ref{subsec:optimization} and Section \ref{subsec:regression} were performed on the FLAME face model, which has yet a different topology (5023 vertices) than the four topologies used in this experiment.

\section{Conclusion}
\label{sec:conclusion}
In this work we propose a novel perceptual shape loss function for monocular 3D face reconstruction. Inspired by a study on the human perception of shape~\cite{Ramachandran1988}, we design a discriminator (critic) network that learns to distinguish `good' (real) face reconstructions from `bad' (fake) ones, by only looking at pairs of input images and gray shaded geometry renders. This is accomplished by creating a dataset of real and fake face reconstructions, and by training the critic network to output higher scores for the real examples than for the fake examples. Once trained, our critic's score, which we refer to as the perceptual shape loss, can be plugged into any optimization or neural network regression framework for evaluating the quality of a given face reconstruction. We demonstrate the ability of our shape loss to successfully improve upon face reconstructions during parameter optimization. We also show that our shape loss is useful when fine-tuning a parameter regression neural network~\cite{deca}. The result is a novel method (PSL) that yields improved results on standard single-image face reconstruction benchmarks~\cite{nowbenchmark,realy}, when compared to the initialization and other state-of-the-art methods that do similar fine-tuning with different perceptual loss terms\cite{emoca,spectre}. Further, as our critic operates only on shaded renders of a 3D face shape, our perceptual shape loss is agnostic to topology of the 3D face mesh, thereby making it suitable for use across a large variety of face models. 

\subsection{Limitations and Future Work}
Our method in its current state provides limited benefits for optimizing or training on side poses, primarily because of the lack of synthetic in-the-wild examples used to train the critic. Currently when using our loss on side poses, we found that the additional landmark term is required to keep the mesh in the correct pose. We however note that this is a limitation of the dataset generation technique \cite{neuralfacedb} and not of our method itself. Furthermore, on some occasions we notice that our method does not fully close the mouth of individuals with closed mouth expressions. We think this might come from the mouth crop that we apply to the ground truth geometries and might be alleviated by cropping less of the inner mouth region away in future work. Even though we followed Ramachandran~\cite{Ramachandran1988} and based our metric on the perception of shading and used gray shaded renders as input to our critic, one might imagine that a more thorough exploration of the light, and material properties used to create the geometry render could lead to improved results. We leave such an exploration to future work. Finally another line of research could explore the design space of critic networks to perhaps obtain local score estimates~\cite{schonfeld2020u} which could provide finer signals for supervision.

\bibliographystyle{eg-alpha-doi}

\bibliography{egbib}
\newpage
\appendix
\subsection{A. Architecture Details}
\begin{figure}[t]
	\center
	\includegraphics[width=0.79\columnwidth]{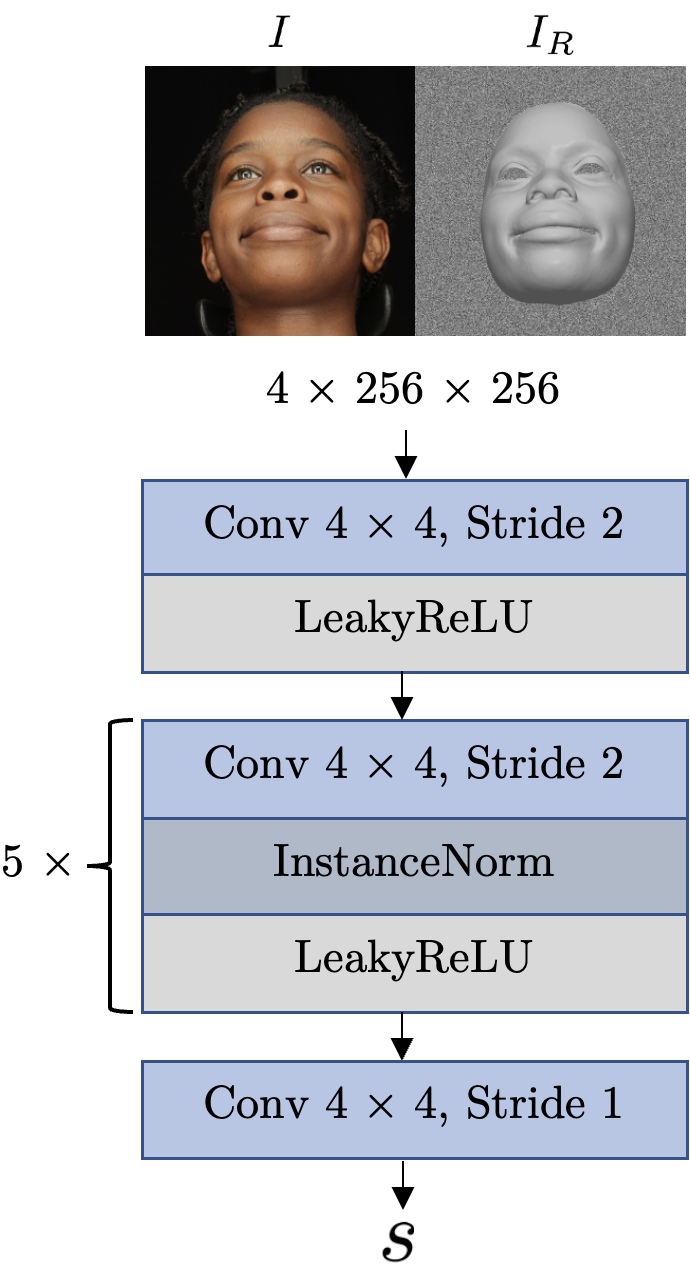}
	\caption{Architecture details for our convolutional network $\mathcal{D}$ ~\cite{wgangp,wgan}. The LeakyReLU activations use a negative slope of $0.2$.}
	\label{fig:architecture_details}
\end{figure}

Figure \ref{fig:architecture_details} shows more details of our convolutional network $\mathcal{D}$~\cite{wgangp,wgan}, which takes the aligned RGB image and gray-shaded render with noise background as a four channel input. Output is a real-valued 1-dimensional scalar $\mathcal{S}$, representing the perceptual shape loss score.

\subsection{B. Implementation Details}

The results for the EMOCA\_V2~\cite{emoca} version that we compare to in the paper are based on their \emph{EMOCA\_v2\_lr\_mse\_20} model checkpoint, which uses the SPECTRE~\cite{spectre} lip reading loss. 

As SPECTRE~\cite{spectre} itself is a video-based method, we create small 10 frame videos consisting out of the same single-image frame for running their reconstruction. We process the video with chunk size 10 and extract the last valid reconstruction from the first chunk.

\subsection{C. Additional Results}

\begin{figure*}[h!]
	\center
	\includegraphics[width=0.82\textwidth]{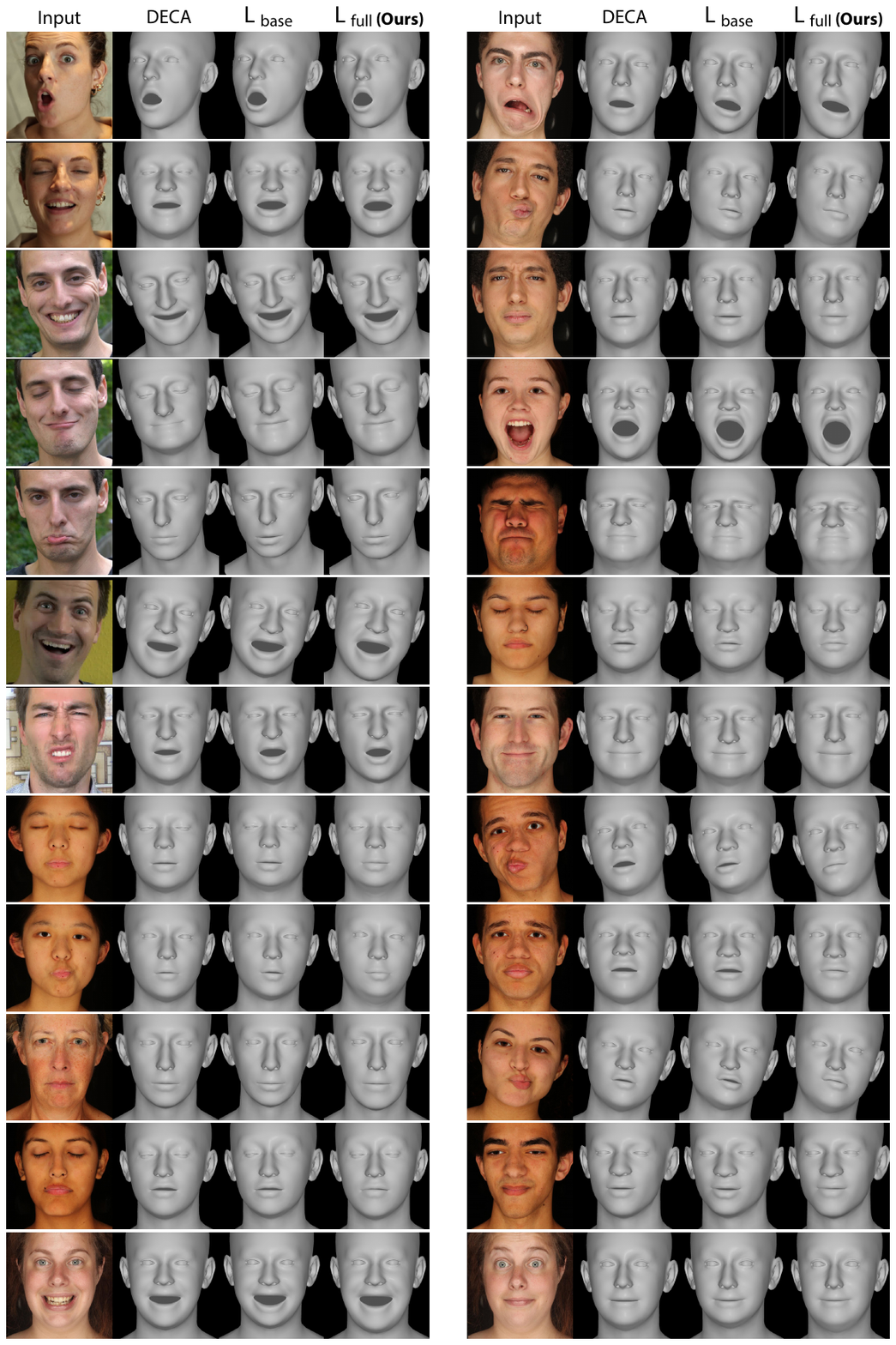}
	\caption{We show additional qualitative optimization results for face reconstructions on in-the-wild data and on our held-out validation set. The optimization is initialized by the DECA~\cite{deca} parameters. We show the results for the traditional losses in the column ($\mathcal{L}_{\text{base}}$) and the results including our proposed perceptual shape loss in column ($\mathcal{L}_{\text{full}}$).}
	\label{fig:more_results}
\end{figure*}

\begin{figure*}[h!]
	\center
	\includegraphics[width=0.8\textwidth]{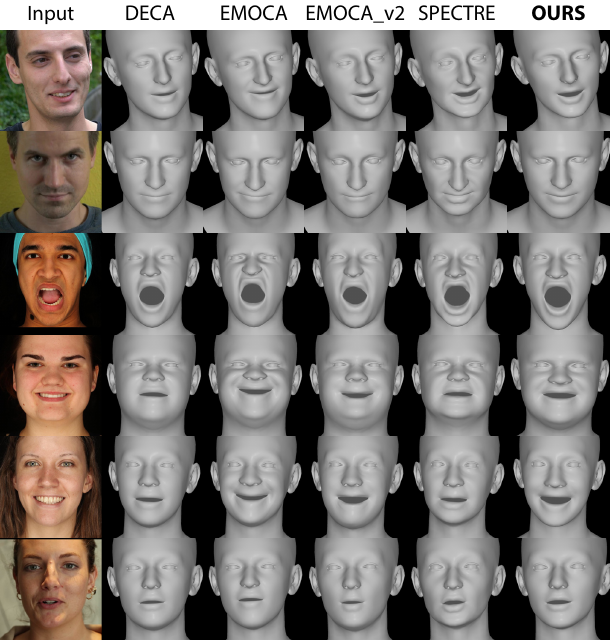}
	\caption{Here we show more qualitative inference-based results for face reconstructions, comparing the initial DECA~\cite{deca} fit with our new method PSL - $\mathcal{L}_{\text{full}}$ and related state-of-the-art methods EMOCA~\cite{emoca}, EMOCA\_V2~\cite{emoca} and SPECTRE~\cite{spectre}.}
	\label{fig:inference-results-sup}
\end{figure*}

We present additional optimization results using our perceptual shape loss in Figure \ref{fig:more_results}. The results are generated following the same optimization schedule as mentioned in the main paper. For all parameter updates we use the AdamW optimizer \cite{adamw} with a learning rate of $0.005$.\\
Additional inference-based qualitative comparisons with related state-of-the-art work are presented in Figure \ref{fig:inference-results-sup}.

\label{sec:supplementary}

\subsection{D. Data of Human Subjects}

For all of the personal data that is shown in the main paper and in this supplementary material, we have obtained the consent of the respective individuals.

\end{document}